\pdfoutput=1

\documentclass[11pt]{article}

\usepackage{acl}
\usepackage{times}
\usepackage{latexsym}
\usepackage{booktabs}
\usepackage{pdfpages}
\usepackage{threeparttable}
\usepackage{hyperref}
\usepackage{float}
\usepackage[T1]{fontenc}

\usepackage[utf8]{inputenc}

\usepackage{microtype}
\usepackage{multirow, array}

\usepackage{inconsolata}

\usepackage{graphicx}

\title{RadBARTsum: Domain Specific Adaption of Denoising Sequence-to-Sequence
Models for Abstractive Radiology Report Summarization }

\author{
 \textbf{Jinge Wu\textsuperscript{1}},
 \textbf{Abul Hasan\textsuperscript{1}},
 \textbf{Honghan Wu\textsuperscript{1,2}}
\\
\\
 {\textsuperscript{1}University College London,
 \textsuperscript{2}University of Glasgow}
\\
  \texttt{\{jinge.wu.20, a.kalam, honghan.wu\}@ucl.ac.uk} \\
}

\begin{document}
\maketitle
\begin{abstract}

Radiology report summarization is a crucial task that can help doctors quickly identify clinically significant findings without the need to review detailed sections of reports. This study proposes RadBARTsum, a domain-specific and ontology facilitated adaptation of the BART model for abstractive radiology report summarization. The approach involves two main steps: 1) re-training the BART model on a large corpus of radiology reports using a novel entity masking strategy to improving biomedical domain knowledge learning, and 2) fine-tuning the model for the summarization task using the ``Findings'' and ``Background'' sections to predict the ``Impression'' section.
Experiments are conducted using different masking strategies. Results show that the re-training process with domain knowledge facilitated masking improves performances consistently across various settings. This work contributes a domain-specific generative language model for radiology report summarization and a method for utilising medical knowledge to realise entity masking language model. The proposed approach demonstrates a promising direction of enhancing the efficiency of language models by deepening its understanding of clinical knowledge in radiology reports.

\end{abstract}

\section{Introduction}




Radiology reports play a crucial role in medical diagnosis and treatment planning. However, their often lengthy and technical nature can pose challenges for both doctors and patients \cite{bosmans2011radiology}. Automatic summarization techniques offer a promising solution, enabling the creation of concise summaries that highlight key findings and expedite clinical workflow. For clinicians, summaries can prioritize critical information within radiology reports, saving valuable time previously spent sifting through extensive text \cite{bosmans2011radiology}. This improved efficiency allows doctors to focus on decision-making and patient care. Additionally, summarization can facilitate the identification of crucial findings and abnormalities within reports, enhancing diagnostic accuracy \cite{zhang2018learning, van2023radadapt}.

Traditionally, radiology reports consist of three sections: Background, Findings, and Impression (as whosn in Table ~\ref{Tab:radio_sample}). The radiologist meticulously details observations in the Findings section before synthesizing these details into a concise Impression, which encapsulates the core clinical takeaways.

Recent advancements in Natural Language Processing (NLP), particularly the emergence of Large Language Models (LLMs) like GPT-3, have revolutionized the field of text summarization \cite{radford2018improving, achiam2023gpt}. However, applying these models to clinical reports remains a challenge due to the inherent complexity of medical terminology and the vast domain-specific knowledge required for accurate summarization \cite{zambrano2023rales}. Furthermore, the sheer size and parameter count of LLMs often render fine-tuning for specialized tasks like radiology report summarization cost-prohibitive \cite{van2023radadapt, zambrano2023rales, wu2022survey}.


Prior research has demonstrated the effectiveness of fine-tuning encoder-decoder based language models such as BART, T5, and PEGASUS on clinical reports \cite{lewis2019bart, zhang2020pegasus, zhang2020pegasus}. However, these models are susceptible to generating factual errors, often introducing nonsensical content or "hallucinations" due to their limited understanding of medical terminology \cite{wu2023knowlab}.


To address these limitations, our study proposes a novel approach that leverages a two-step fine-tuning process based on BART model. First, we implement a customized entity Masked Language Modeling (MLM) strategy to enhance the model's understanding of medical terminology. Subsequently, we further fine-tune the model on a text summarization task specifically tailored to radiology reports for optimal performance. This approach holds promise for generating accurate and informative summaries of radiology reports, ultimately benefiting both clinicians and patients in the medical decision-making process.

\begin{table}[htp]
\centering
\setlength{\tabcolsep}{1.5mm}
\begin{tabular}{|p{7.3cm}|} 
\hline
 \textbf{Background}: History of lung cancer. \\  
 \textbf{Findings}: Lung volumes are low. There may be mild pulmonary vascular congestion. The heart size is borderline enlarged. The mediastinal and hilar contours are relatively unremarkable. Innumerable nodules are demonstrated in both lungs, more  pronounced in the left upper and lower lung fields compatible with metastatic disease. No new focal consolidation, pleural effusion or pneumothorax is seen, with chronic elevation of right hemidiaphragm again seen. The patient is status post right lower lobectomy. Rib deformities within the right hemithorax is compatible with prior postsurgical changes. \\
\hline
\textbf{Impression}: Innumerable pulmonary metastases. Possible mild pulmonary vascular congestion. Low lung volumes.\\
\hline
\end{tabular}
    \caption{An example of radiology report. The text summarization task is to predict the "Impression" based on the information from "Findings" and "Background".}
\label{Tab:radio_sample}
\end{table}

In summary, our contributions include:
\begin{enumerate}
\item We introduce an entity masking strategy to enhance medical knowledge acquisition during pre-training. By masking medical terms in the masked language modeling phase, we encourage the model to focus on domain-specific knowledge, leading to improved understanding of medical concepts and their relationships.
\item We propose RadBARTsum, a BART-based model trained with entity masking for radiology report summarization. Our experimental results show its effectiveness compared to the baseline, highlighting its potential to streamline clinical workflows and improve medical documentation efficiency.
\end{enumerate}


\section{Related Work}

\subsection{Sequence-to-Sequence Models}
The Sequence-to-sequence (Seq2Seq) models, introduced by \citet{sutskever2014sequence}, initially transform an input sequence into a fixed-dimensional vector utilizing a neural network such as an LSTM or Transformer \cite{vaswani2017attention}, referred to as the encoder. Subsequently, another network of the same type is employed to decode the target sequence from this vector \cite{sutskever2014sequence}. This encoder-decoder framework has proven effective across various NLP tasks, including, but not limited to, machine translation, document summarization, and text generation \cite{zhang2020pegasus, lewis2019bart}. 
In this study, we concentrate on Transformer-based Seq2Seq models. Notable examples of such architectures include BERT \cite{devlin2018bert}, RoBERTa \cite{liu2019roberta}, GPT \cite{radford2018improving}, T5 \cite{raffel2020exploring} and BART \cite{lewis2019bart}. These models have demonstrated state-of-the part performances across spectrum of NLP tasks. These models vary in size, prtraining methodologies, and the tasks they are applied to. BART \cite{lewis2019bart} employs an arbitrary noising function to corrupt the input text and subsequently learns to reconstruct the original text. This training paradigm has demonstrated remarkable effectiveness, particularly in the task of text summarization, showcasing BART's ability to generate coherent and concise summaries from noisy input data. 

\subsection{Masked Language Modelling}
Masked language modeling (MLM) is an important and efficient task for pre-training language models where a portion of the input text is masked or hidden and the model is required to predict the masked tokens \cite{devlin2018bert}. MLM has been shown to be effective at learning contextual representations of language and has been used to achieve state-of-the-art results on a variety of NLP downstream tasks \cite{devlin2018bert}. 
Specifically, in the original BERT model 15\% of tokens are selected from the input text to replace them with a special [MASK] token \cite{devlin2018bert}.
There are different masking strategies applied in various work. For example, XLNet \cite{yang2019xlnet} model is trained using a variant of the MLM task called permutation language modeling, which randomly permut the order of the tokens in the input text and training the model to predict the original order of the tokens \cite{yang2019xlnet}. Conversely, ALBERT \cite{lan2019albert} and Longformer \cite{beltagy2020longformer} develop sentence and document masking strategies, respectively (i.e. mask the entire sentence or document). RoBERTa apply the masking strategy called whole word masking, by replacing entire words with a special [MASK] token rather than individual tokens within a word \cite{liu2019roberta}. Finally, the masking strategy for BART \cite{lewis2019bart} model involves replacing a portion of the tokens with six types of noises, including token masking, token deletion, text infilling, sentence permutation and document rotation. The probability of a token being masked or deleted is set to 50\% in the BART model \cite{lewis2019bart}. Our work extends this masking strategy with selective masking on clinical entities.
\subsection{Radiology Report Summarization}

Early work by \citet{dredze2008automatic} laid the groundwork for this task by introducing a system that utilized NLP techniques to automatically generate summaries of radiology reports. Since then, the field has witnessed significant advancements. \citet{chen2018radiology} presented a method leveraging attention-based neural networks for radiology report summarization. \citet{zhang2018learning} explored the problem of generating radiology impressions through summarizing textual findings. Their approach, employing an augmented pointer-generator model, achieved high agreement with human-generated summaries. \citet{macavaney2019ontology} further improved summarization quality by incorporating an ontology-aware pointer-generator model. While \citet{li2019hybrid} and \citet{liu2019clinically} explored generating textual descriptions from medical images using reinforcement learning (RL), research efforts directed towards improving pre-training methods for language models in this domain remained limited.
In 2021, \citet{abacha2021overview} introduced the Radiology Report Summarization track in the MEDIQA 2021 shared task. This initiative demonstrated the effectiveness of fine-tuned BART and PEGASUS models, achieving superior performance compared to previous methods. \citet{hu2022graph}) adopted a similar approach, utilizing a BERT-based model. More recent advancements include the CLIN-T5 model by \citet{van2023radadapt}, which leverages the T5 architecture and lightweight adaptation on the MIMIC-CXR dataset. Additionally, \citet{adams2024speer} showcased the effectiveness of entity extraction to guide open-source LLMs for improved summarization. Furthermore, independent studies by \citet{liu2023exploring} and \citet{ma2023impressiongpt} demonstrated that few-shot prompting with GPT-4 outperformed all other LLMs in this task. However, adapting such massive models to clinical settings presents challenges, including computational resource limitations, data privacy concerns, and ethical considerations.

\section{Method}

This study intends to conduct further training on BART model \cite{lewis2019bart}, a denoising autoencoder for pre-training sequence-to-sequence models. It is a combination of BERT \cite{devlin2018bert} and GPT \cite{radford2018improving} model, which inherites the state-of-the-art (SOTA) performance of both bidirectional encoder and left-to-right decoder models. Moreover, being constructed on Seq2Seq Transformers architecture, BART is an excellent candidate for abstractive summarization, as it has the ability to generate novel text and paraphrase the input text. It does this by using an encoder-decoder architecture, where the encoder processes the input text and generates a representation of it, and the decoder generates the summary based on this representation. The model is trained on a large dataset of text-summary pairs, and is able to learn how to generate summaries that are coherent and semantically similar to the original text.

One of the key advantages of BART is its ability to handle long input sequences and generate summaries that are much shorter than the original text. This is achieved through the use of the Transformer architecture, which allows the model to efficiently attend to different parts of the input text and selectively incorporate information from different parts of the sequence.

As mentioned before, the orginal pre-training of BART model involves in MLM task with six noises, which is an enhancement of BERT model with single noise on masking strategy. This makes the model less likely to learn biased information. However, this will makes it harder to learn domain specific knowledge when it happens to knowledge intensive clinical notes. Therefore, we add one more training process (i.e. re-training) with medical entity masked language modelling (MLM) to help improve the understanding of domain knowledge.

\subsection{Medical Entity Masked Language Modelling (MLM)}
In the MLM task, selective masking is a variant of the masking strategy where only a subset of the tokens in the input sequence are masked. This is in contrast to the standard MLM approach, where a random selection of tokens are masked.

Selective masking can be useful because it allows the model to focus on specific types of tokens that are important for the downstream task. In our case, the medical entities are selected as masked tokens, this will help inject biomedical domain knowledge into the language model during the re-training process. To achieve this, we applied SemEHR \cite{wu2018semehr}, a clinical NLP tool for medical entity detection. With a given clinical dataset, it will return with all entities found from Unified Medical Language System (UMLS\footnote{https://www.nlm.nih.gov/research/umls/index.html}), which is the most comprehensive medical vocabulary system.

Then these entities will be replaced with a [MASK] token in the input sentences. The model will try to predict the masked tokens during training. As sometimes the entities are phrases instead of a single word, such as "eosinophilic pneumonia" or "tunneled central venous catheter". We conducted both experiments on the phrase-level entity masking and word-level entity masking. For phrase-level entity masking, the whole phrase is replaced with a [MASK] token. For word-level entity masking, the phrases are first splited into a set of word list and then used for tokenization. It is worth noting that during the entities used for entity masking are added into the vacabulary for tokenization to make sure the tokenizer can precisely found the entities. This will lead to an increase in the dimension of embedding layer during the training of phrase-level entity masking and word-level entity masking. 

\begin{figure}[h!]
  \centering
  \includegraphics[width=0.48\textwidth]{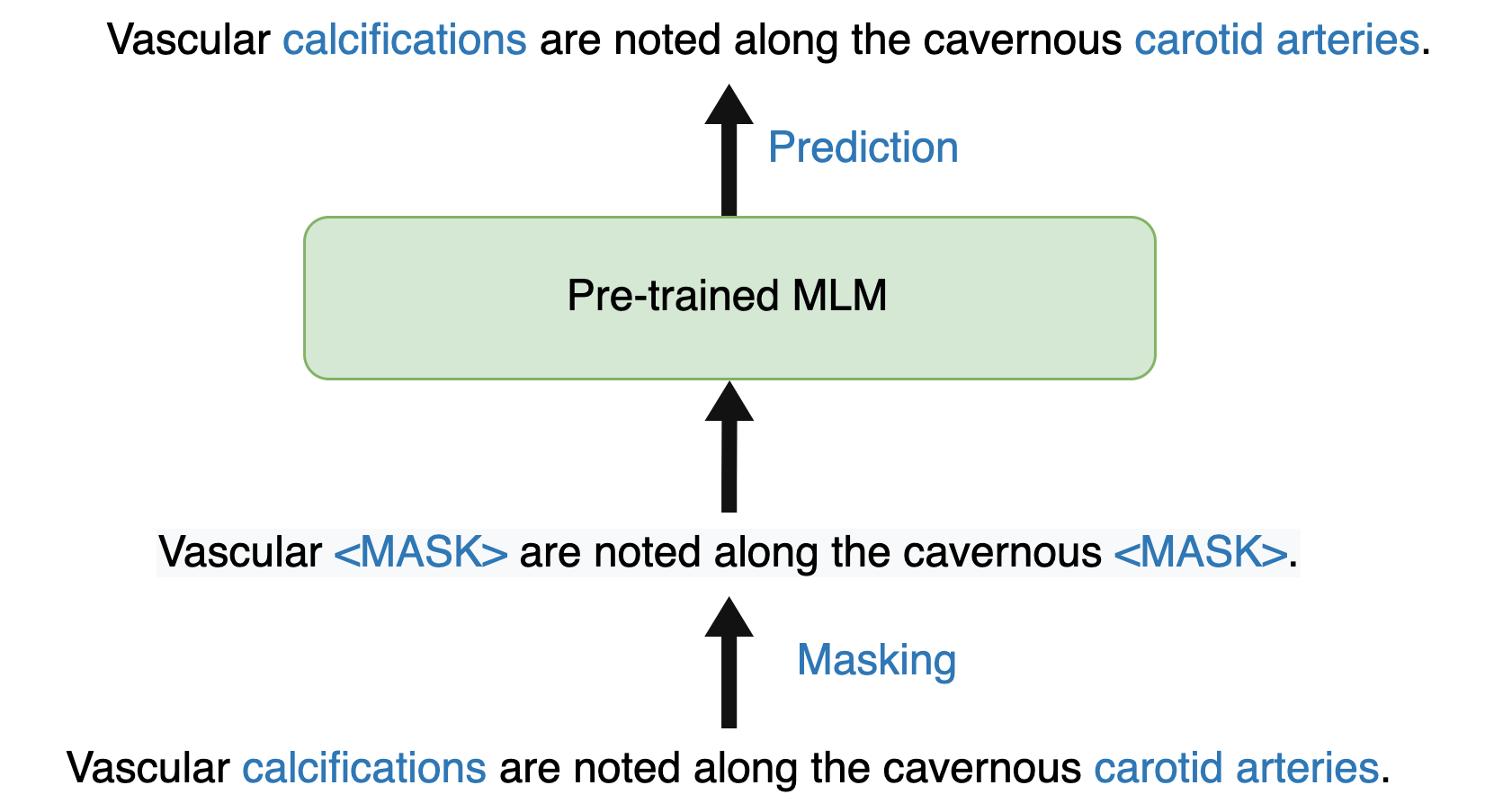}
  \caption{Architecture of the medical entity masking.}
  \label{fig:SMask}
\end{figure} 

\subsection{Fine-tuning}

In this stage, we fine-tune the model from the previous training into text summarization. The input data for this is the "FINGDINGS" section in the radiology and the model will predict the "IMPRESSION" section for summarizaiton. As similar to the re-training process, the tokenizer dimension will be expanded in the phrase-level entity masking model and word-level entity masking model.

\begin{figure*}[h!]
  \centering
  \includegraphics[width=0.9\textwidth]{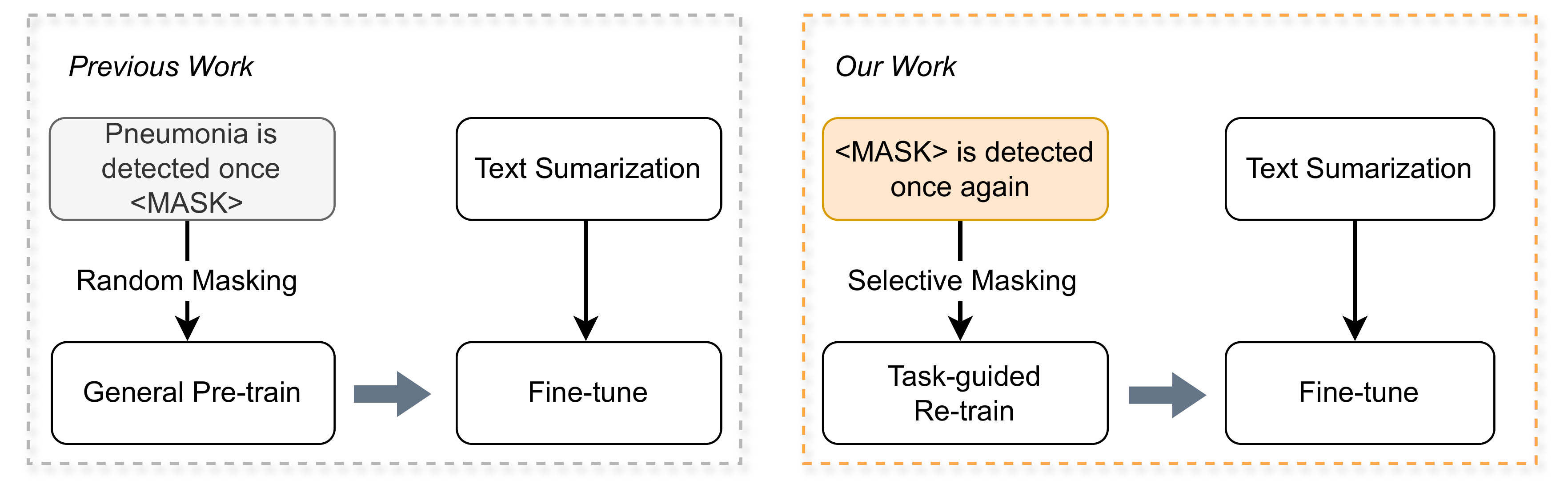}
  \caption{Our proposed work compared with the previous work.}
  \label{fig:label2}
\end{figure*} 

\subsection{Data Preparation}



For our study, we use the radiology reports from MIMIC-III to retrain our models. MIMIC-III is a database that contains free-text EHRs for over 40,000 patients who receives critical care at the Beth Israel Deaconess Medical Center between 2001 and 2012. To ensure that our retraining and fine-tuning data are separate and do not have overlapping data (since fine-tuning data used MIMIC-CXR, which is a subset of MIMIC-IV dataset), we use the MIMIC-III Clinical Database CareVue subset\footnote{https://physionet.org/content/mimic3-carevue/1.4/}, which does not include any patients from MIMIC-CXR.

After retraining on the MIMIC-III dataset, we further fine-tuned the model using MIMIC-CXR radiology reports \cite{johnson2019mimic}. For the test dataset, we use two datasets: 1) a subset of reports from MIMIC-CXR; 2) reports Stanford Health Care Syetem. The test set was partially drawn from a hospital system different from the training set in order to test the model's generalizability. There are no overlapping samples between the train, validation, and test sets for both the retraining and fine-tuning data. Table \ref{Tab:data} shows the splits for each.

\begin{table}[h!]
\small
\centering
\setlength{\tabcolsep}{1.5mm}
\begin{threeparttable}
\begin{tabular}{lllll} 
\toprule
 \multirow{2}{*}{}& \multicolumn{2}{l}{\textbf{Re-train}}   & \multicolumn{2}{l}{\textbf{Fine-tune}}  \\
 \textbf{Split}  & \textbf{\#} & \textbf{Source} & \textbf{\#} & \textbf{Source}  \\
\midrule
Train  &  148,451 & MIMIC-III\tnote{1} & 91,344 & MIMIC-CXR \\
Val I  &  20,000 & MIMIC-III & 2,000 & MIMIC-CXR \\
Test I  &  5,000 &MIMIC-III & 2,000 & MIMIC-CXR \\
Test II  &   &  & 300 & Stanford\tnote{2} \\
\bottomrule
\end{tabular}
\begin{tablenotes}    
        \footnotesize             
        \item[1] MIMIC-III refers to the MIMIC-III Clinical Database CareVue subset, including all kinds of radiology reports (not only chest X-ray)
        \item[2] Stanford refers to the Stanford Health Care system chest X-ray reports
        
      \end{tablenotes}           
    \end{threeparttable} 
    \caption{Data statistics for different step including re-training and fine-tuning.}
\label{Tab:data}
\end{table}

\subsubsection{Data preprocessing}

For each radiology report, it contains multiple sections to record the patient's information. This study only looks at the FINGDINGS and IMPRESSION for text summarization task. Thus for the fine-tuning process, it is necessary to remove all other sections in the reports. For retraining, in order to have as much useful information as possible as well as avoid unnecessary information, we consider include "MEDICAL CONDITION", "REASON FOR THIS EXAMINATION" and all sections in the "FINAL REPORT" (this may include "HISTORY", "INDICATION", "COMPARISON", "FINDINGS", "IMPRESSION", etc.). After extracting these sections, a further cleaning is performed including removing illegal characters, bullet numbers, date time, and heading names. Another series of processing is conducted before input the mode. This may involve tokenizing the text, padding or truncating sequences to a fixed length, and converting the text to numerical vectors using the pre-trained model's vocabulary.

\subsection{Evaluation}

The re-training models are evaluated with cross entropy loss, accuracy and perplexity for the MLM task. In terms of perplexity, it is a common measure used to evaluate the performance of language models. It is defined as the exponentiated average of the cross-entropy loss over all examples in the test set. Mathematically, perplexity can be calculated as follows:

\begin{equation}
    PPL = 2^{H(W)}=2^{-\frac{1}{N} log_2 P(w_1,s_2,...,w_N)}
\label{eq:ppl}
\end{equation}
where $H(W)$ is the cross-entropy given a sequence of words $W$ of length $N$ and a trained language model $P$. A lower perplexity indicates a better-performing model.

We then evaluate the performance of summarization task on the test set using during fine-tuning process using ROUGE score\cite{lin2004rouge} and Bert score \cite{zhang2019bertscore}. For ROUGE score, we calculate the ROUGE-L, which measures the overlap of common subsequences between the two summaries. Given a reference summary $X$ with the length of $m$ and its predicted summary $Y$ with the length of $n$, the recall and precision of ROUGE-L can be calculated as follows:

\begin{equation}
    R_{ROUGE-L}=\frac{LCS(X,Y)}{m} 
\label{eq:rouge1}
\end{equation}
\begin{equation}
    P_{ROUGE-L}=\frac{LCS(X,Y)}{n}
\label{eq:rouge2}
\end{equation}
where $LCS$ is the maximum length of the Longest Common Sequence. The F1-score of ROUGE-L is a measure of the balance between precision and recall for the LCS overlap between the predicted summary and the reference summary. It can be calculated using the following formula:

Then the F1-score of ROUGE-L can be calculated as:
\begin{equation} 
\label{eq:equ_lm1} 
   F1-score = 2 \times \frac{R\times P}{R + P}
\end{equation}

ROUGE is a metric that compares the words in an automatically generated summary with those in a reference summary, and counts how many words match exactly. However, this can be a problem when evaluating abstractive summaries, because a good abstractive summary may use different words with the same meaning as those in the reference summary. To address this issue, we used a different evaluation metric called BERTscore \cite{zhang2019bertscore}. BERTscore measures the similarity between the words in the automatically generated summary and the reference summary by comparing the contextual embeddings of those words, which are generated using a pre-trained BERT model. This metric is useful for detecting paraphrases, and it takes into account the context and word order of the words being compared. It has also been found to be highly correlated with human evaluations.



\begin{table*}[htp]
\centering
\begin{tabular}{lllllllll}
\hline
Model & \multicolumn{2}{l}{MIMIC-CXR}           & \multicolumn{2}{l}{Stanford}    &    \\
                    & RL               & BS        & RL               & BS                  & PT/FT \\ \hline
BART                & 39.9747          & 70.5431          & 35.7879          & 68.1451             & FT    \\
BioBART             & 38.3795          & 70.2176          & 35.2960          & 68.0261          &            FT    \\ \hline
BART(Random Mask)         & 40.4708          & 70.5038          & 37.6574          & 69.7164                & PT+FT \\
BART(Entity Mask(phrase)) & 41.8476          & 70.9747          & 37.8788          & 69.9310                & PT+FT \\
BART(Entity Mask(word))   & \textbf{42.6696} & \textbf{71.5513} & \textbf{38.2225} & \textbf{70.8574}  & PT+FT \\ \hline
\end{tabular}
  \caption{\label{tab:results}
   Model performances on MIMIC-CXR and Stanford radiology datasets. RL - ROUGE-L score; BS - BERT Score; PT - pre-trained; FT - fine tuned. }
\end{table*}

\section{Experiments and Results}

\subsection{Experiment Setup}

\begin{table}[htp]
\centering
\footnotesize
\begin{tabular}{llll}
\hline
  Model                  & ACC             & Loss            & PPL           \\ \hline
BART(Random Mask)        &  0.8816          & 0.5223          & 1.8251          \\
BART(Entity Mask(phrase)) & 0.8882          & 0.4861          & 1.5198           \\
BART(Entity Mask(word))   &  \textbf{0.8952} & \textbf{0.4317} & \textbf{1.3824}  \\ \hline
\end{tabular}
  \caption{\label{tab:retrain}
   Comparison of retraining effectiveness between different masking language modelling strategies. ACC - accuracy; PPL - perplexity.}
\end{table}

This study investigates the impact of two masking strategies on the performance of a pre-trained BART model: random masking and selective (medical entity) masking. Random masking follows the same strategy employed in the BERT MLM task, where 80\% of tokens are replaced with the [MASK] token, 10\% are replaced with random words, and the remaining 10\% remain unchanged. In contrast, selective masking focuses on medical entities. We mask medical entities extracted from SemEHR. Two entity lists are explored: a word-level list containing 4,950 entities and a phrase-level list with 14,483 entities. The original BART model has a vocabulary size of 50,265 tokens, including special tokens.  When the medical entity lists are incorporated into the tokenizer for both word-level and phrase-level models, the vocabulary size increases to 55,215 and 64,748 tokens, respectively.  Furthermore, to assess the influence of the entity lists themselves, we conduct experiments using both the word-level and phrase-level entity lists within the random masking strategy. We also try different proportion of masked entities to find a optimal solution for MLM strategy.


In the fine-tuning stage, all the models that have undergone the re-training process are fine-tuned specifically for the text summarization task. To provide a comprehensive comparison, we also fine-tune two baseline models: the BART base model and the BioBART model \cite{yuan2022biobart}. It is worth noting that these baseline models are fine-tuned without the re-training process. BioBART is a domain-specific generative language model that adapts the BART architecture to the biomedical domain by pre-training on extensive PubMed corpora. This pre-training enables BioBART to capture the nuances and intricacies of biomedical language. By comparing the performance of our domain-adapted model, which has undergone re-training, with that of BioBART, we can gain valuable insights into the effectiveness of our approach in the context of biomedical text summarization. This comparison allows us to assess whether our re-training process yields improvements over the existing domain-specific model and helps us understand the extent to which our model can generate accurate and coherent summaries in the biomedical domain.

\textbf{Re-training Setup} \quad We re-train BART base model on MIMIC-III Clinical Database CareVue subset with the training split in Table \ref{Tab:data}. The initial learning rate is 5e-5 with a "linear" lr\_scheduler\_type and warmup\_steps of 500 steps. The maximum epochs used for training is set to 20 with batch size of 4. Other hyper-parameters are set as their default values. 

\textbf{Fine-tuning Setup} \quad We fine-tune models using MIMIC-CXR for training and validation, Indiana dataset for validation and test, Standford dataset for test, see in Table \ref{Tab:data}. The initial learning rate is 2e-5. The maximum epochs used for training is set to 20 with batch size of 16. The maximum length for input data is 1024. The maximum length for output is 128. The beam size is set to 5, and no\_repeat\_ngram\_size is set to 2. Other hyper-parameters are set as their default values. 

\subsection{Results}
\begin{table*}[htp]
  \centering
  \begin{tabular}{p{2.5cm}p{11.5cm}}
    \hline
    \textbf{Input}      & There are small bilateral residual pneumothoraces. visualized non-contrast appearance of the soft tissues of the neck is otherwise unremarkable. the overall alignment of the cervical spine is preserved without listhesis. there is no evidence of fracture, dislocation, or bony destruction. there is slight prominence of the prevertebral soft tissues up to 5 mm in diameter, although there is no evidence of osseous injury. although not ideal for visualization of the thecal contents, the visualized intrathecal contents appear unremarkable by ct.           \\
     \hline
    \textbf{Ground Truth}    & 1. no evidence of fracture or alignment abnormality. 2. mild prominence of the prevertebral soft tissues, which are not convex however to suggest a hematoma. please correlate clinically. 3. bilateral pneumothoraces.         \\
    \hline
    \textbf{BART}    & 1. no evidence of vascular abnormality. 2. neck soft tissues appear normal on non-contrast imaging.         \\
        \hline
    \textbf{BioBART}    & 1. unremarkable soft tissues of the neck 2. no fractures, dislocations, or bony destruction.  3. normal mra of the neck       \\
        \hline
    \textbf{Ours}    & 1. no significant abnormalities in the soft tissues of the neck, cervical spine alignment, or visualized intrathecal contents. 2. small bilateral residual pneumothoraces 3. slight prominence of prevertebral soft tissues, no osseous injury.       \\
    \hline
  \end{tabular}
  \caption{\label{tab:qual_analysis}
   Generated ``Impressions'' from different models for a sample report.
  }
\end{table*}

Table \ref{tab:results} presents the overall performance of all models. The results demonstrate that the proposed entity masking approach effectively improves performance in terms of both ROUGE and BERTScore on the MIMIC-CXR and Stanford datasets, highlighting its efficacy in enhancing the model's understanding of medical text. It is noteworthy that both word-level and phrase-level masking strategies contribute to performance gains, with word-level masking consistently achieving the best results across all evaluation metrics.

Table \ref{tab:retrain} evaluates the effectiveness of MLM training. The results align with Table \ref{tab:results}, where medical masking for both word-level and phrase-level gain better performance than random masking, exhibiting the higher accuracy and the lower loss and perplexity. These findings underscore the importance of focusing on medical entities during the masking process, as it enables the model to better capture the domain-specific knowledge essential for accurate text generation.

\begin{figure}[h!]
  \centering
  \includegraphics[width=0.45\textwidth,trim={80 0 500 250},clip]{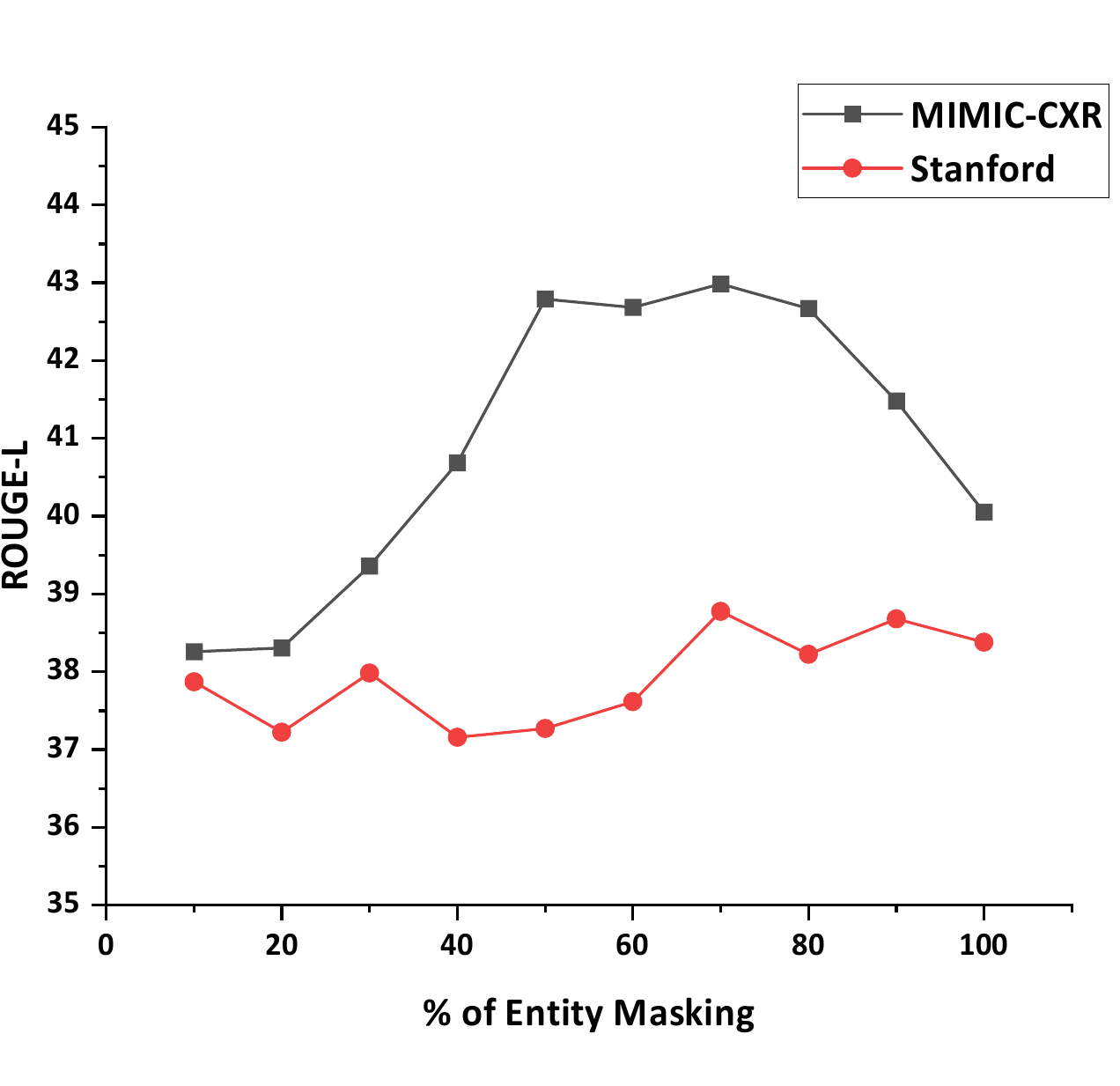}
  \caption{\label{tab:rank}
   The effect of the proportion of medical entity masking on the summarization task performance.}
\end{figure}

To further investigate the impact of the proportion of medical entities during masking, an ablation study is conducted, as shown in Figure \ref{tab:rank}. The table presents the results of masking words with varying proportions of medical entities. For both datasets, the ROUGE-L scores generally increase as the percentage of entity masking rises. The MIMIC-CXR dataset consistently achieves higher scores compared to the Stanford dataset across all entity masking percentages.
The graph exhibits a steady upward trend for both datasets until around 80\% entity masking, after which the MIMIC-CXR dataset continues to improve, while the Stanford dataset's performance slightly declines.

In summary, the proposed entity masking approach, which focuses on masking medical entities, proves to be an effective strategy for improving the performance of text generation models in the medical domain. The ablation study further confirms the importance of prioritizing medical entities during the masking process, as it enables the model to better capture domain-specific knowledge and generate more accurate and coherent medical text.

\paragraph{Qualitative Analysis.} 
Table \ref{tab:qual_analysis} provides a qualitative analysis of the models' performance, with given input and the summaries. Compared to BART and BioBART, our model appears to provide the most comprehensive and precise summary compared to the ground truth. While the BioBART and BART models perform well in certain aspects, their summaries suffer from omissions of important information or lack of detailed findings. For BioBART, while it correctly notes ``unremarkable soft tissues of the neck", ``no fractures, dislocations, or bony destruction", and ``normal MRI of the neck", it lacks any description of the key finding of bilateral residual pneumothoraces. This may result in an incomplete summary of the key information. For BART, although it accurately identifies ``no evidence of vascular abnormality" and ``normal appearance of the neck soft tissues on non-contrast imaging", it does not mention the presence of bilateral residual pneumothoraces. Also, BART model's summary seems overly simplistic and limited. It does not address details such as the preservation of cervical spine alignment and the slight prominence of the prevertebral soft tissues.

\section{Discussion}


This work presents a novel method for infusing domain knowledge into open-domain language models for the task of radiology report summarization. We achieve this by first pre-training a MLM model that masks medical entities within the training data. Subsequently, the model is fine-tuned on the specific task of summarizing radiology reports. Our experiments explore various approaches to incorporating these clinical entities, including phrase-level and word-level injection into the tokenizer. Additionally, we investigate different masking strategies to enhance the LLM's understanding of medical terminology. By masking medical entities, we aim to direct the model's attention towards acquiring domain-specific knowledge during the fine-tuning process. The effectiveness of this approach is demonstrated by the results presented in the preceding section.

However, there are several areas for improvement in future work. First, developing a better tokenization method is crucial. In our case, we did not train an adapted tokenizer from scratch; instead, we input the medical tokens as extra vocabulary in the existing tokenizer. This approach may lead to suboptimal representations of the contextual information associated with these terms. Therefore, investigating a more sophisticated tokenizer should be a priority in future research.

During our experiments, we also observe that the main errors stem from hallucinations, where the model generates nonsensical or factually incorrect sentences. In some cases, the model repeatedly generates the same sentences, which are quite frequent in the training datasets. These findings highlight the need for further improvement in the model's ability to generate diverse and accurate content.

Moreover, optimizing the fine-tuning process is essential for maximizing the model's performance on downstream tasks. This study primarily focuses on the re-training process without significant enhancements to the fine-tuning stage, which may result in the underperformance of the MLM when evaluated on downstream tasks. Future work should also include human evaluations of the test data to provide a more comprehensive assessment of the model's performance and to identify areas for further improvement.

In addition to the aforementioned improvements, future research could explore the integration of external knowledge bases or ontologies to further enhance the model's understanding of domain-specific concepts and relationships. This approach could potentially mitigate the issue of hallucinations and improve the model's ability to generate accurate and coherent summaries.

\section{Conclusion}

In this study, we develop methods for radiology report summarization by proposing an entity masking approach for re-training a domain-specific generative language model. Our model is built upon BART, a denoising sequence-to-sequence model, which is first re-trained with medical entity masking language modelling and then fine-tuned for the specific downstream task of text summarization. The results demonstrate that the proposed re-training process improves models' performances, and the word-level entity masking strategy is superior consistently across different evaluation settings. Despite the limitations and challenges identified, such as the need for a more sophisticated tokenizer, hallucination mitigation, and fine-tuning optimization, this study presents a promising approach for injecting domain knowledge into open-domain language models, paving the way for more accurate and efficient radiology report summarization.

\bibliography{custom}

\appendix



\end{document}